\documentclass[letterpaper, 10 pt, journal, twoside]{IEEEtran}
\usepackage{cite}
\usepackage{amsmath,amssymb,amsfonts}
\usepackage{algorithmic}
\usepackage{graphicx}
\usepackage{subcaption}
\usepackage[hidelinks]{hyperref}
\usepackage{textcomp}
\usepackage{booktabs}
\usepackage{xcolor}
\usepackage{svg}
\usepackage{verbatim}
\usepackage{pifont}
\usepackage{multirow}
\definecolor{darkgreen}{rgb}{0.0, 0.5, 0.0}
\usepackage{footnote}
\makesavenoteenv{table}
\usepackage{graphicx}
\graphicspath{{./contents/figures/}}

\newcommand{\method}{{GATAR}}

\begin{document}

\title{Graph-based Decentralized Task Allocation for Multi-Robot Target Localization}

\markboth{IEEE Robotics and Automation Letters. Preprint Version. Accepted September, 2024}
{Peng \MakeLowercase{\textit{et al.}}: Graph-based Decentralized Task Allocation forMulti-Robot Target Localization} 

\author{Juntong Peng, Hrishikesh Viswanath and Aniket Bera%
\thanks{Manuscript received: May 16, 2024; Revised August 1, 2024; Accepted September 16, 2024. This paper was recommended for publication by Editor J. Kober upon evaluation of the Associate Editor and Reviewers' comments.}
\thanks{The authors are with the Department of Computer Science, Purdue University, USA,
        {\tt\small \{juntong, hviswan, aniketbera\}@purdue.edu}}%
}

\maketitle

\newcommand{\Note}[1]{{\color{blue} \bf \small [NOTE: #1]}}

\begin{abstract}

We introduce a new graph neural operator-based approach for task allocation in a system of heterogeneous robots composed of Unmanned Ground Vehicles (UGVs) and Unmanned Aerial Vehicles (UAVs). The proposed model, \texttt{\method}, or \textbf{G}raph \textbf{A}ttention \textbf{T}ask \textbf{A}llocato\textbf{R} aggregates information from neighbors in the multi-robot system, with the aim of achieving globally optimal target localization. 
Being decentralized, our method is highly robust and adaptable to situations where the number of robots and the number of tasks may change over time. We also propose a heterogeneity-aware preprocessing technique to model the heterogeneity of the system. 
The experimental results demonstrate the effectiveness and scalability of the proposed approach in a range of simulated scenarios generated by varying the number of UGVs and UAVs and the number and location of the targets. We show that a single model can handle a heterogeneous robot team with a number of robots ranging between 2 and 12 while outperforming the baseline architectures. 
\end{abstract}

\section{Introduction}

\begin{figure}[!htp]
    \centering
    \includegraphics[width=0.36\textwidth]{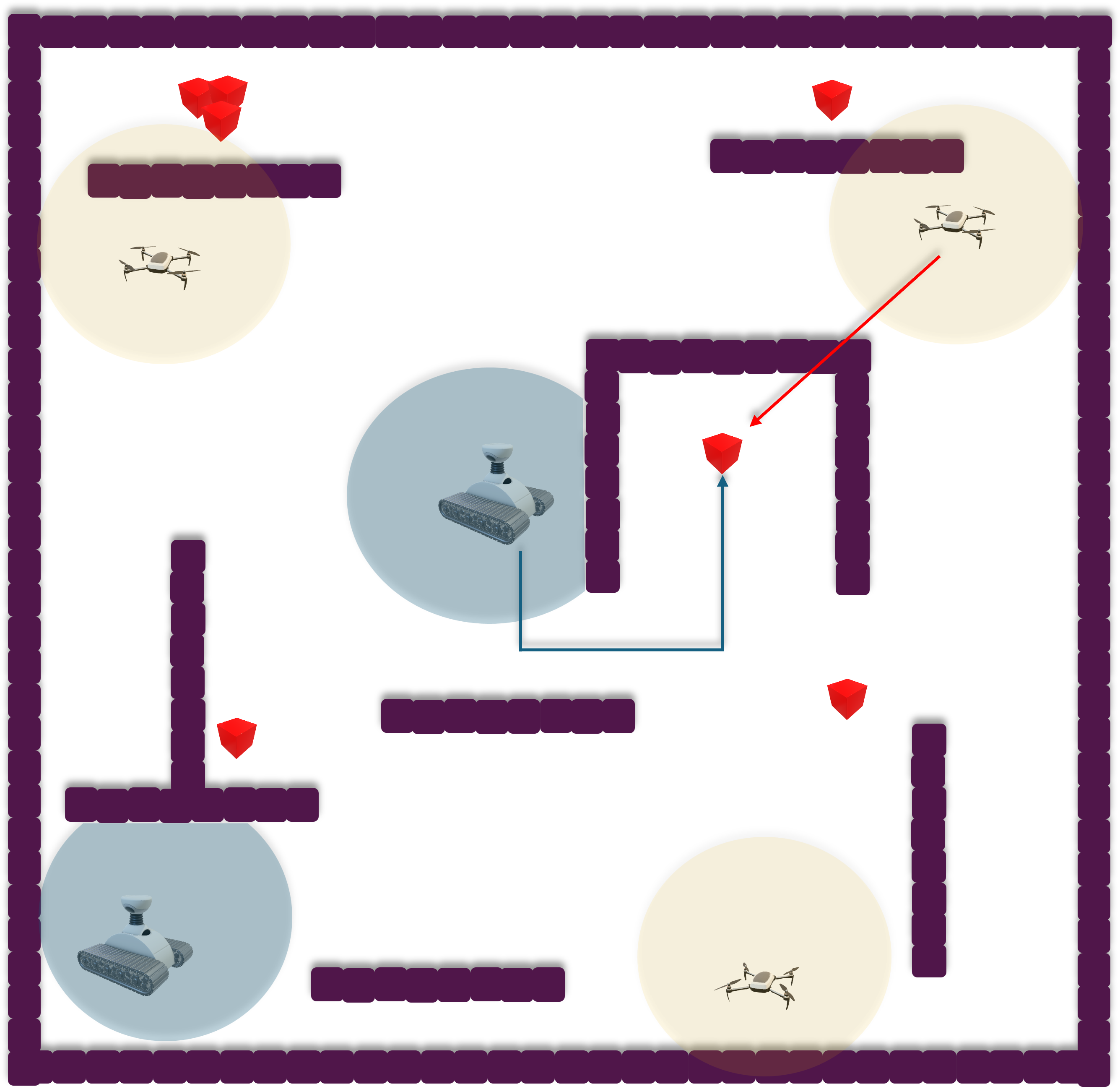}
    \caption{\textbf{System Overview}: This figure highlights that by utilizing comparative advantages, a collaborative MRS can perform efficient target localization. The UAV can fly over the ground obstacle, compared to the UGV, which navigates around it, making it more efficient for UAV to be assigned the target within the U-shaped obstacle}
    \vspace{-15pt}
    \label{fig:abstract}
\end{figure}

\IEEEPARstart{I}{n} recent years, multi-robot systems have gained significant attention due to their potential in a myriad of applications, from surveillance and reconnaissance to exploration and disaster response \cite{cortes2017coordinated, arai2002advances, gautam2012review, rizk2019cooperative, parker2007distributed}. The ability of multi-robot teams to distribute tasks, cover larger areas, and collaboratively process information makes them ideal for scenarios that are demanding, hazardous, or expansive, particularly in the realm of target localization—where pinpointing the precise location of a target in a given environment is of paramount importance. The collaboration of multiple robots can improve the efficiency and the speed of the task. However, there are several challenges in multi-robot systems, such as coordinating diverse robot types, such as Unmanned Ground Vehicles (UGVs) and Unmanned Aerial Vehicles (UAVs), ensuring real-time communication, and effectively processing vast amounts of collected data. Figure \ref{fig:abstract} highlights how robots with different physical capabilities can collaboratively navigate to their assigned targets and achieve globally optimal performance. 

Conventional centralized methodologies encounter limitations in scalability in extensive environments, besides being susceptible to single points of failure \cite{bertin2009distributed}.  Furthermore, centralized methodologies typically require communication characterized by high robustness and minimal latency. However, in cases where the terrain is particularly challenging or the geographical expanse is vast, it is difficult to provide such performance guarantees. Consequently, decentralized approaches have increasingly become more popular due to their robustness against communication interruptions and member failures, as well as their ability to scale easily \cite{parker2009cooperative, de2006decentralized, liu2014decentralized, zhang2018fully}.

Facilitating communication between multiple robots is important for collaboration, as it ensures that every robot is aware of what the other robots know. Graphs are an efficient data structure for communicating between multiple robots, and they can vary in topological structures. They also inherently enable the aggregation of information across multiple robots. Given the dynamic nature of mission scenarios, the use of a graph allows for dynamic scaling with the number of robots, thereby ensuring the robustness of the system as the number of robots changes.

In complex and unknown environments, heterogeneous multi-robot systems are preferred to homogeneous ones because different types of mobile robots offer different operational advantages and operate under varying resource constraints\cite{tiwari2018estimating,tiwari2019unified}. For example, when comparing UGVs with UAVs, UGVs have a relatively long operational range and can carry heavier payloads, while UAVs, despite limited payload-carrying ability, are relatively faster and more agile. Therefore, a heterogeneous robot team is better suited for collaborative work in an uncharted environment\cite{prorok2021beyond}.

There have been attempts at utilizing heterogeneous multi-robot collaborations in unknown terrains~\cite{pimenta2008sensing,santos2018coverage,kim2022coverage,notomista2019optimal,malencia2022adaptive,debord2018trajectory,emamm2020Adaptive,emam2021data,mayya2021resilient}. A recently proposed general reinforcement learning method, Heterogeneous Graph Neural Network Proximal Policy Optimization is capable of dealing with behavioral and physical heterogeneity, and allows each robot to learn a different set of parameters based on their physical attributes\cite{hetGPPO}. Another multi-agent reinforcement learning (MARL) method proposed a class-wise policy to deal with heterogeneity between agent classes\cite{seraj2022learning}. These two methods \cite{hetGPPO,seraj2022learning} achieve satisfactory performance. However, their network architectures and weight need to be modified when new robots are added to the environment, thereby affecting the scalability of the system. 

In this paper, we propose a graph neural operator-based decentralized method for heterogeneous multi-robot target localization in unknown environments. Our major contributions are as follows:
\begin{itemize}
    \item \textbf{Novel graph neural operator based \texttt{\method}\footnote{\href{https://github.com/IDEAS-Lab-Purdue/GATAR}{https://github.com/IDEAS-Lab-Purdue/GATAR}}}: The model uses a V-Cycle algorithm based graph attention network, defined over a radius graph, to aggregate information from neighbors, which is more sensitive to information from a relatively long range and more adaptable to variations in the number of neighboring nodes. 
    \item \textbf{Heterogeneity-aware preprocessing}: We model the observed local information while taking into consideration the robots' heterogeneity in mobility. Different types of robots can now share a single aggregation and execution model, making it possible to add or remove different species or robots on the fly, making the network scalable.
    \item \textbf{Two stage task localization}: To prevent robots' current tasks from being repeatedly altered due to the detection of new targets in the vicinity, we split the target localization problem into two stages: first, the task allocation, and second, path planning conditioned on the assigned task. The path-planning stage can be effectively addressed using numerous well-established algorithms.
\end{itemize}

\section{Related Works}
In this section, we survey numerous decentralized target localization approaches in Sec.~\ref{subsec:target-loc} followed by a survey of current deployment strategies for heterogeneous multi-robot systems in Sec.~\ref{subsec:mrs}. We conclude our literature review with a survey of current advancements in Graph Neural Networks (GNNs) in Sec.~\ref{subsec:gnn}, which will serve as a building block for our proposed multi-robot task allocation model.
\vspace{-5pt}
\subsection{Decentralized Target Localization}\label{subsec:target-loc}

Various multi-robot approaches have been explored in the past, ranging from conventional stereo geometry-based approaches \cite{li2023toward, thayer2001distributed, zou2019collaborative, zhang2022cvids, olfati2006flocking} to the newer machine learning-based approaches \cite{zhura2023neuroswarm, li2022online, arel2010reinforcement, shen2021graphcomm, liu2022multi, yu2023learning, li2021fatigue}. In their work, Dias et al. use a probability-based triangulation method, using stereo geometry to localize the position of an object in 3D space from monocular observations of the robots \cite{dias2015decentralized}. Tallamraju et al. define the problem of multi-robot target tracking and obstacle avoidance as a convex optimization task subject to constraints defined by robot dynamics \cite{tallamraju2018decentralized}. However, their work focuses on a homogenous system of MAVs or Micro Aerial Vehicles. 
While these works introduce various ways to model multi-robot systems, there is a lot of potential to improve the scalability, capture longer-range information, and build a system that supports robots entering and leaving the environment. 
\vspace{-10pt}
\subsection{Heterogeneous Multi-Robot System (MRS)}\label{subsec:mrs}
A heterogeneous multi-robot system refers to a setup wherein the robots have different physical properties and functional properties. An example of a system with physically diverse robots is the one with both UGVs and UAVs. A functionally diverse heterogeneous system could be a system of robots where some are reconnaissance, some are combat, and some are rescue robots. Darintsev et al. define high-level policies to implement a control system for a system of functionally diverse robots \cite{darintsev2019methods}. However, a centralized coordinator is needed. Ponda et al. design a consensus-based distributed auction algorithm to perform task allocation in heterogeneous settings \cite{ponda2010decentralized}. Their approach handles time constraints, fuel costs, and robots' physical properties. Similarly, Sariel-Talay et al. propose a generic distributed framework \textit{DEMiR-CF} for efficient task allocation in multi-robot environments \cite{sariel2011generic}. Pujol et al. present a Tractable Higher Order Potential (THOP) based model for heterogeneous robot collaboration in urban rescue operations \cite{pujol2015efficient}. 
\vspace{-10pt}
\subsection{Graph Neural Network}\label{subsec:gnn}
Among the works that used graph-based models for task allocation, Oh et al. designed a particle swarm optimization framework based on graph theory for optimal task allocation where directed acyclic graphs represent constraints \cite{oh2016pso}. 
Li et al. propose graph neural networks for decentralized multi-robot path planning, where the action is predicted from the current state, goal, and observable map \cite{li2020graph}. Their method, however, utilizes a single goal state in the input. Zhou et al. explore the use of graph neural networks for decentralized task allocation \cite{zhou2022graph}. They use local communications between robots to determine the action that each robot takes. They show that this approach works better than decentralized greedy approaches in coarse-grained planning. 
HetNet is a graph neural network-based model for implementing communication in heterogeneous collaborative tasks \cite{seraj2022learning}. However, this model relies on class-wise interpreters and exploration of the environment, which may weaken the generalizability to some extent.
Li et al. posit a key-query-based architecture called Message-Aware Graph Attention neTwork (MAGAT), which uses spatial attention to determine the relative importance of different features in the message received from the neighbors \cite{li2021message}. 
Zhang et al. propose GResNet, a residual network style architecture incorporating GNN layers for improving the learnability of the model \cite{zhang2019gresnet}. 
Among the recent works that propose reinforcement learning-based strategies, Agarwal et al. \cite{agarwal2019learning} propose a scalable approach with an agent-entity graph, where the edges connect agents to entities. They propose a selective attention mechanism to process information from neighboring agents. Jiang et al. \cite{jiang2018graph} and Khan et al. \cite{khan2020graph} propose an RL model with a graph convolution network (GCN) for multi-agent interactions. The model is adapted to a dynamically changing environment. Their approach, however, doesn't include target localization tasks. infoMARL, proposed by Nayak et al. \cite{nayak2023scalable}, is a multi-agent goal allocation model where the individual agent's observations are concatenated and fed to an actor model to generate the actions. Hu et al. \cite{hu2023graph} propose a MARL architecture that uses a GNN to parameterize Gaussian policies in order to predict actions for the agents.

While most Graph Convolution approaches aggregate information between close neighbors, it is important to have global information when the task allocation is decentralized. To address this, we propose a new graph-based architecture and incorporate elements from the graph neural operator \cite{li2020multipole} to facilitate long-range interactions between robots.

\section{Problem Formulation}

\textbf{Decentralized task allocation for target localization} is defined as the assignment of one target from an observable set of targets to each robot based on its current observation, the proximity to the target, and the environment information aggregated from its local neighbors, defined by a radius graph. The location of the target is used for collision-free trajectory prediction. We extend the environment defined in \cite{zhou2022graph} to make it more complex by ensuring the distribution of obstacles within the environment is non-uniform in terms of concentration and inter-obstacle spacing. Furthermore, the system of robots within the environment is heterogeneous, comprising both aerial and ground robots. The problem is formulated as a sequential task allocation problem. The goal is to optimize the target allocation strategy in order to either localize more targets within a set number of steps or to reduce the time required to localize all the targets.

Consider a system with N robots $A=\{a_0, a_1 \ldots a_N\}$ in an environment represented by $E \in \mathbb{R}^2$. There are M static targets $S =\{s_0, s_1, \ldots s_M\}$. 

\textbf{Assumptions} 1) Every robot knows its own location, $p_i$, in the currently observed environment; 2) The global expanse of the environment and the location of the targets are initially unknown to all the robots 3) The robots can observe the environment around them within a sensing range $r_{sense}$, defined as the field-of-view and there could be obstacles, which occlude the observation for UGVs; 4) The robots within a communication range $r_{comm}$,  can exchange information. This is defined as the neighborhood radius in a radius-based graph; 5) The mobility of UAVs isn't affected by ground obstacles, and the UAVs have a birds-eye view of the targets. 6) The obstacles in the environment are not uniformly distributed. 7) The environment is discretized into a grid world. Thus, the grid world can have obstacle-rich regions and obstacle-free regions.

\textbf{Individual Observation and Preprocessing:}
We can express the behavior of the $i$-th robot in the system as follows:
\begin{subequations}
\vspace{-5pt}
    \begin{align}
    o_{i,t} &= \Phi_{E}(p_i, r_{comm_i})\label{eq1}\\
    x_i  &= \Phi_{P}(o_{i,t}, m_i)\label{eq2}
    \end{align}
\end{subequations}
In the above equations, $o_{i,t} \subseteq E$ refers to the observable environment of robot $i$ at timestep $t$, which is a function of the position $p$ of the robot and the physical sensing capability $r$. 
The encoded features of robot $i$, represented as $x_i$ are a function of the observable environment of the robot and the mobility $m_i$ of the robot, representing the physical attributes distinguishing UGVs from UAVs.

\textbf{Decentralized Message Passing and Graph-based Aggregation:}
The aggregated feature map of the robots within a neighborhood is denoted as follows. 
\begin{subequations}
    \begin{align}
    X_{t} &=[ x_1,  x_2,\hdots x_N]^T\label{eq3}\\
    \mathcal{N}_i &= \Phi_C(r_{comm_i},X_{t})\label{eq4}
    \end{align}
\end{subequations}
In the above equations, $x_i$ denotes the features of robot $i$ while $\mathcal{N}_i$ is the aggregated features of the neighbors of robot $i$, which is a subset of the complete set $X_{t}$, depending on the communication range $r_{comm}$ of robot $i$. 
\begin{subequations}
    \begin{align}
    [S_t \mathcal{N}_t]_{if} &= \sum_{j : a_i \epsilon \mathcal{N}_i} [S_t]_{ij}[\mathcal{N}_t]_{jf} = \sum_{j: a_i \epsilon \mathcal{N}_i} S_t^{ij}n_t^{jf}\label{eq5}\\
    \mathcal{C}(\mathcal{N}, S) &= \sum_{k=0}^K S^k \mathcal{N} H_k\label{eq6}
    \end{align}
    \label{eq3all}
\end{subequations}

In the above equations, Eq.~\eqref{eq5} represents the linear combination of the features of the neighboring robots. $S_t$ is the graph shift operator, which is an adjacency matrix to describe the radius-based communication graph. Eq.~\eqref{eq6} represents the graph convolution operation where $S$ is the adjacency matrix at that time, $K$ represents the maximum hops of the neighbors included in the collaboration, $\mathcal{N}$ is the node features and $H_k$ represents the weights.
\vspace{-5pt}
\section{Architecture}\label{arch}

\begin{figure*}[!h]
    \centering
    \vspace*{5pt}
\includegraphics[width=0.85\textwidth]{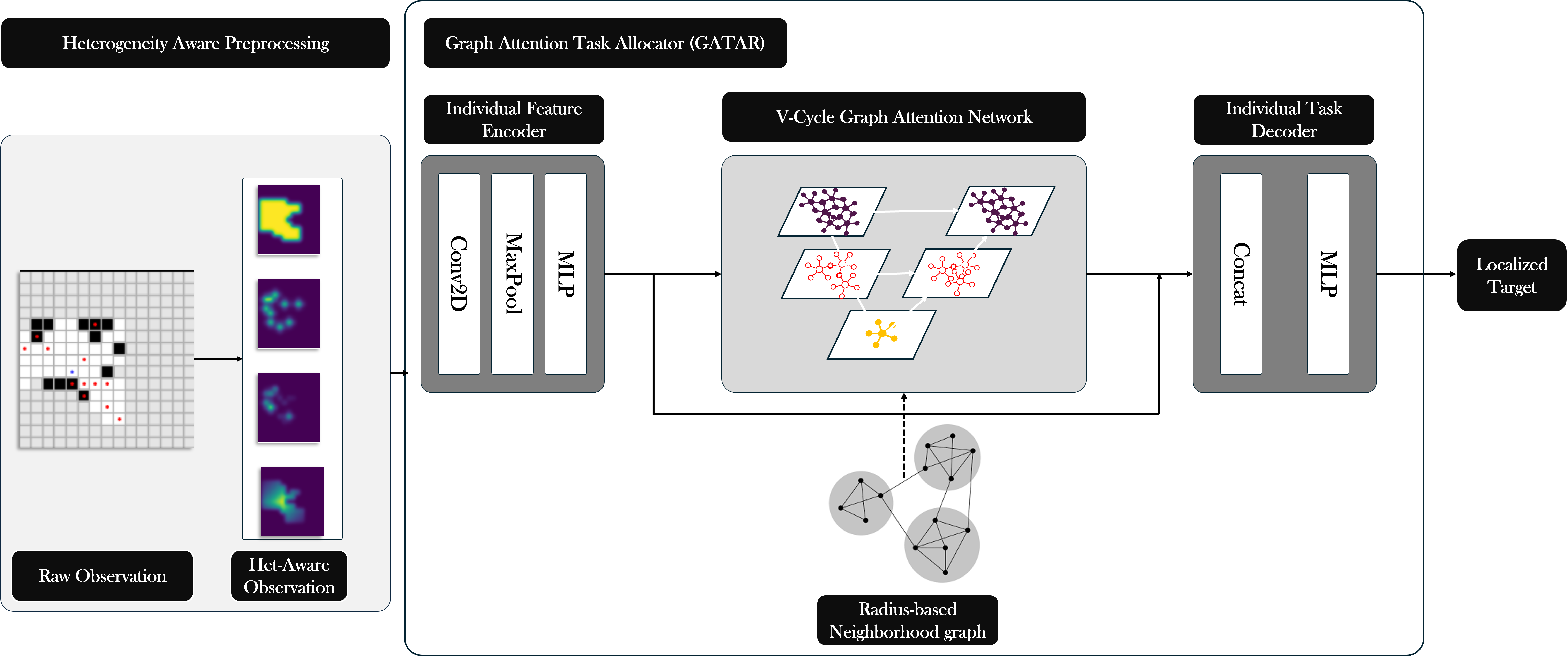}
    \caption{\textbf{GATAR Architecture and Het-Aware preprocessing:} The model takes as input the observable environment, targets, obstacles and robots as input and performs information aggregation through the V-cycle algorithm before generating the x-y coordinates corresponding to the location of the target, which can be used for path planning}
    \vspace{-15pt}
    \label{fig:method}
\end{figure*}

In this section, we describe our decentralized task allocation model that allows the Multi-Robot System (MRS) to operate collaboratively to localize initially unknown targets in a locally observable environment. 

As shown in Figure \ref{fig:method}, our method consists of two parts: The first component is the ``heterogeneity-aware observation preprocessing," wherein each robot builds feature maps based on its observation and mobility.  This local information is broadcast to all neighbors within the communication range. The second component is the graph neural operator based  ``\method" or \textbf{G}raph \textbf{A}ttention \textbf{T}ask \textbf{A}llocato\textbf{R}, which uses a radius graph to aggregate information from all the robots inside the neighborhood radius and capture long-range information through the v-cycle algorithm, with linear complexity. This aggregated information is used to assign targets to each of the robots in the environment. The following subsections discuss the two components in detail. 
\vspace{-5pt}
\subsection{Heterogeneity-aware Observation Preprocessing}

We propose a universal representation to capture the physical characteristics of robots in a heterogeneous system, by encoding these characteristics into the feature embedding as given in Equation ~\eqref{eq2}. 

The heterogeneity is defined by two properties: 
\begin{enumerate}
    \item \textbf{Perception ability}: This is defined as whether the line-of-sight is occluded by ground obstacles.
    \item \textbf{Mobility}: This represents whether the robot is impeded by ground obstacles.
\end{enumerate}

We design a four-channel spatial feature map to model heterogeneity. The first channel is an \textbf{Reward map}, which determines the reward of the possible actions for a robot in the sensed regions. The reward is determined as a function of the length of the shortest path from the robot's current position to the furthest points in the sensed area. 
The heuristic for the reward $R$ that a robot $i$ gets by moving to a region $l$ within the currently observed environment $E_i$ is denoted as
\begin{equation}
    R_{i, l} = \frac{v_i}{d_{i, l} + \epsilon} \quad \forall_{i \in \mathbf{R}} \forall_{l \in E_i}
\end{equation}
where $v_i$ is the velocity of the robot and $d_{i,l}$ is the shortest path from the current position to the location $l$, and $\epsilon$ is a small positive constant. The heuristic incentivizes the robot to prioritize shorter paths over longer paths. The second channel is a \textbf{Gaussian target map}. Each target in this map is represented as a target-centered spatial Gaussian distribution to prioritize the region around the targets as well. For every target $\tau$ within the set of targets $T$ in the observed region $l \in E_i$ of the robot $i$, at a distance $(d^x_{l,\tau}, d^y_{l, \tau})$, the Gaussian target map is defined as
\begin{equation}
    \mathbb{G}_{l,\tau} = exp(-(\frac{d^x_{l,\tau}}{\sigma^x_i}^2+\frac{d^y_{l,\tau}}{\sigma_i^y})) \quad \forall_{l \in E_i} \forall_{\tau \in T}
\end{equation}
The third channel is an \textbf{Observed obstacle map} in the form of a Boolean matrix, representing the location of the known obstacles. The last channel is the pixel-wise product of \textbf{Reward map} and \textbf{Gaussian target map}. This channel works as a handcrafted feature that indicates the spatial preference of each robot, where the regions around the target, within close proximity of the robot, have the highest preference.
\vspace{-10pt}
\subsection{Graph Attention Task Allocator (\method)}

\texttt{\method} is a graph neural operator-based architecture that aggregates information between robots through message passing and uses this information for task allocation. 

Message passing is a local short-range operation that aggregates information between neighbors within a radius. This operation can be defined as 
\begin{equation}
    h_{u}^{k+1} = \mathbb{U}(h_{u}^k, \mathbb{A}(h_x^k, \forall_{x \in \mathcal{N}(u)})
\end{equation}
In the above equation, $\mathbb{A}$ and $\mathbb{U}$ refer to aggregate and update operations, respectively. $h_u^{k}$ is the node feature $u$ at iteration $k$ and $\mathcal{N}(u)$ is the neighborhood of $u$. 

In order to allow for scaling to an arbitrary number of robots, which may be distributed non-uniformly in the environment, it is necessary to aggregate long-range information to provide a global understanding of the environment. However, in such cases, having a single large connected graph would increase the time complexity of the network. 
To learn long-range information with linear time complexity, multi-hop neighbor aggregation is done by adapting the V-cycle algorithm proposed in \cite{li2020multipole}, which breaks the graph into several levels based on the range of interaction. 

Using multiple layers of graph neural networks to aggregate information from multihop neighbors suffers from over-smoothing problems, in which nodes lose variations after information aggregation. This will necessarily lead to the assignment of the same task to all the neighbors within a radius. If close neighbors are assigned the same task, the target localization becomes very inefficient. Inspired by ResNet\cite{He_2016_CVPR} and the Multipole Graph Kernel Network\cite{li2020multipole}, we integrate the V-cycle structure to address the mentioned issue. This structure also allows the model to partially reuse parameters when capturing long-range information. For instance, information aggregated from more than 1 hop neighbors all share the parameters of the first layer in a downward pass. Equation \ref{eq:gno4co} mathematically describes this structure with $L$ layers:
\begin{subequations}
\label{eq:gno4co}
\begin{align}
    \check {v}_{l+1} &= \varphi(K_{l+1,l}\check{v}_{l,\pi}), l \in [0,L-1]\label{subeq:down}\\
    \hat{v}_l &= \varphi(K_{l,l}\check{v}_{l, \pi}) +  \varphi(K_{l,l-1}\hat{v}_{l-1, \pi}), l \in [1,L]\label{subeq:up}
\end{align}

\end{subequations}
where Eq.~\eqref{subeq:down} represents the downward pass and Eq.~\eqref{subeq:up} represents the upward pass ($\varphi$ here represents an arbitary learnable function). $\check{v}_l$ and $\hat{v}_{l-1}$, respectively, denotes the $l^{th}$ layer's input features, and therefore $\check{v}_0$ is the input of the features to the V-cycle structure and $\hat{v}_{L}$ is the aggregated features of that structure. $\check{v}_{l,\pi}$ or $\hat{v}_{l, \pi}$ denotes the $l^{th}$ layer's output features, regarding the learnable parameters $\pi$ and $K_{l,l}$ represent the decomposed graphs at different layers. 

For each layer, a graph attention network from \cite{velivckovic2017graph} is adapted to deal with a varying number of neighbors and redundancies in the neighborhood information, which has been shown to be effective in MAGAT~\cite{li2021message}. Eq.~\eqref{eq:gat} mathematically shows the multi-head graph attention network we used:
\begin{subequations}
\label{eq:gat}
\begin{align}
    \alpha_{ij}&=\frac{\exp(\rho(e_{ij}))}{\Sigma_{k\in {\mathcal{N}_i}}\exp(\rho(e_{ik}))}\label{subeq:alpha}\\
    v_{l,k}&= \parallel^{K}_{k=1}\sigma(\Sigma_{j\in \mathcal{N}_i}\alpha_{ij}^kW^kv_l)\parallel
\end{align}
\end{subequations}
where, $\rho$ represents the \texttt{LeakyReLU} activation and $\alpha_{ij}$ is the attention co-efficient. With the attention coefficient calculated in Eq.~\eqref{subeq:alpha}, the neighbors are weighted and the aggregated information is also weighted accordingly. As shown in Fig. \ref{fig:method}, the model uses a sequence of CNNs and MLPs initially, which is residually connected to the decoder, also composed of CNNs and MLPs. The output of the network is the target position assigned to the robot. 
\section{Training}
\textbf{Dataset generation and implementation details}\label{subsec:ds}:
We synthetically generate an obstacle-filled 2D grid world and randomly place robots and targets within the 15$\times$15 grid. We experimented with several configurations of UAVs and UGVs, starting with 1 UAV and 1 UGV initially and increasing to 6 UAVs and 6 UGVs for a total of 12 robots in the scene. We vary the number of initial targets between 20 and 30 for different robot configurations. To train the model, we generate 2000 maps and up to 1000 configurations of robots and targets within those 2000 maps. Each map contains a specific obstacle distribution, and the density of obstacles is varied; each configuration contains a unique target distribution and initial places for the robots. The grid world here can be considered as the discretization of a real-world scene at arbitrary resolutions, as the model's output consists of x-y coordinates for the target (a sample for discretization is shown in Fig.\ref{fig:realworld}). We use an expert algorithm to generate the paths to targets, using \texttt{tcod} library. The obstacles increase the cost of the paths for UGVs but have no impact on UAVs. A subset of these possible paths is used to train the model to allocate the target for each robot. The process occurs until there are no more targets to be assigned to the robots. Typically, the training set for one robot team configuration contains 1.5 million samples, and the validating set contains 400,000 samples. For testing set, the number is 150,000 samples.

\textbf{Training phase:} At higher resolutions of grid world, the computation complexity of task allocation increases, with the exhaustive search having a time complexity of $\mathcal{O}(n!)$. To efficiently sample enough examples for training, we use a centralized greedy algorithm as the expert algorithm, with the aim of obtaining a near-optimal solution. In this setting, however, targets that are not seen by any robots are excluded from allocation. This setting is based on the fact that if a target is not in the field of view of any of the robots, then by no means can the entire multi-robot system know of the existence of that target. The dataset is divided into a training set (80\%) and a validating set (20\%). During inference, we use a different set of maps and configurations.

The model is trained in a supervised learning paradigm, using Mean Square Error (MSE) as the loss function. We perform training and inference on a 32-core A100-40GB GPU node. The loss is defined as 
\begin{equation}
    L(\theta; i) = \parallel \mathcal{G}_\theta(E, i) - \tau_i \parallel_2^2
\end{equation}
where, $\mathcal{G}_\theta$ is the model, which takes the gridworld $E$ as input and predicts the target for robot $i$; $\tau_i$ is the expected target assigned to robot $i$. 

\vspace{-5pt}


\section{Empiricial Evaluation}

We perform several experiments on datasets generated by the expert algorithm. We discuss these in Sec.\ref{subsec:quan}. Furthermore, we run the model on grid worlds defined on real-world map snapshots captured from Google Maps, which we discuss in Sec.\ref{subsec:qual}. 

\vspace{-10pt}

\subsection{Evaluation metrics}

As our task is configured to predict a location that approximates the target position, the primary evaluation metric we employ is the distance between the predicted location and the location generated by the expert algorithm. To better monitor the performance of the model, we evaluate the average distance between predicted targets \& the actual targets (\texttt{dist-avg}), the median distance between predicted \& the actual targets across all samples (\texttt{dist-50}) and the 90 percentile distance across all samples (\texttt{dist-90}). The median distance quantifies the performance of the models in normal cases, while the 90 percentile distance shows how the models deal with the toughest cases, and the average distance shows the overall performance.

\vspace{-10pt}
\section{Results and Ablations}\label{subsec:quan}

\subsection{Results on generated grid world dataset}

\begin{figure*}[ht]
    \centering

    \begin{subfigure}[b]{0.31\textwidth}
        \centering
        \vspace*{6pt}
        \includegraphics[width=\textwidth]{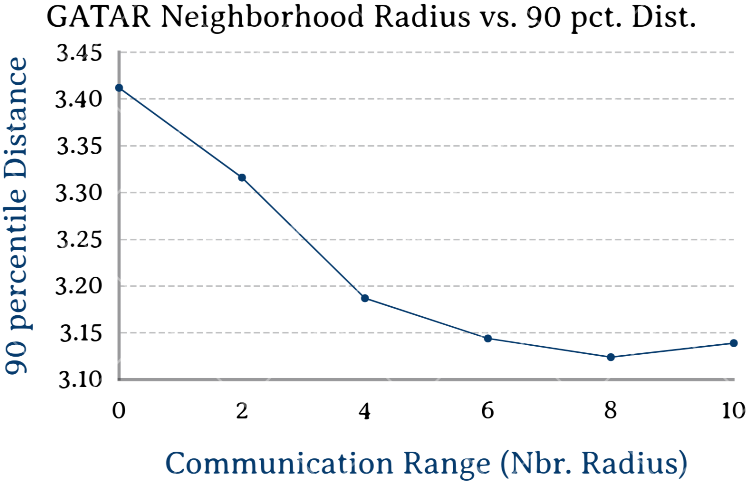}
        \caption{}
        \label{fig:comm_range90}
    \end{subfigure}
    \begin{subfigure}[b]{0.31\textwidth}
        \centering
        \vspace*{6pt}
        \includegraphics[width=\textwidth]{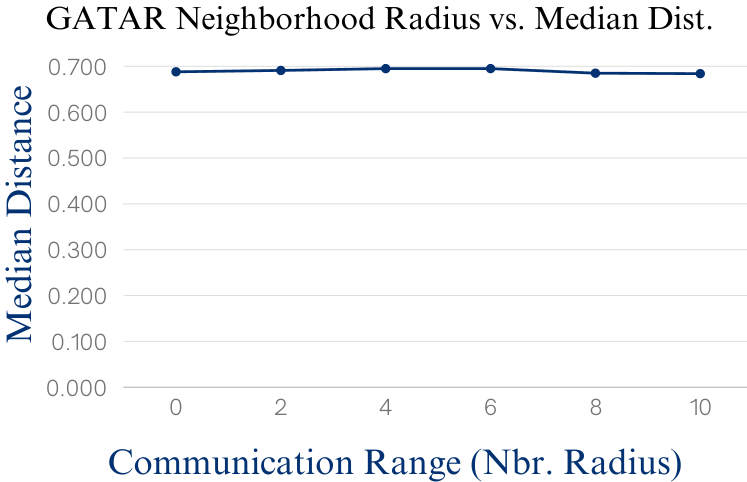}
        \caption{}
        \label{fig:comm_range50}
    \end{subfigure}
    \begin{subfigure}[b]{0.31\textwidth}
    \centering
    \vspace*{6pt}
    \includegraphics[width=\textwidth]{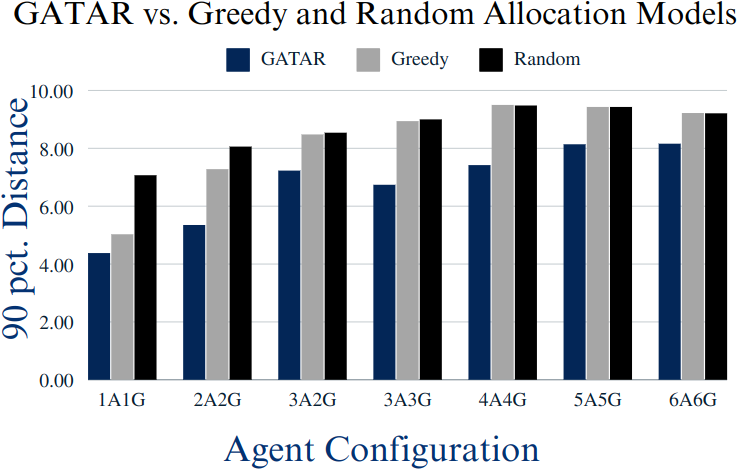}
    \caption{}
    \label{fig:teamsize}
    \end{subfigure}
    \caption{(a) 90th percentile distance, (b) Median distance from the predicted targets to the targets from expert algorithm when the communication range goes up, (c) 90th percentile distance of \method  ~against 
 Greedy algorithm and Random task allocation}
 \vspace{-15pt}
    \label{fig:comm_range}
\end{figure*}

\textbf{Effect of Communication Range}: In our problem setting, the range of communication between robots is constrained by the neighborhood radius. A small communication range means that the robots can only communicate with the closest neighbors, and they need to make decisions, relying mostly on ego-centric observations since the observations of close neighbors almost coincide with each other, while a large communication range means that each robot is able to obtain information from a large number of robots and obtain a near-global understanding of the environment. 
We tested our model against different communication ranges to show the effectiveness of information aggregation between robots. As shown in Fig. \ref{fig:comm_range90}, the 90th percentile distance decreases as the communication range expands, implying that dense collaboration helps the robots deal with corner cases in which the optimal target cannot be decided without aggregating the other robots' knowledge Meanwhile, as shown in Fig. \ref{fig:comm_range50}, when communication ranges or the radius increase from 0 to 10 grid cells, the median distance between the target and the predicted location remains below 1 grid cell, indicating that ego-centric observations suffice when the targets are within the robot's field of view.

\begin{table}[h]
    \centering
    \resizebox{8cm}{!}{
    \begin{tabular}{ccccccc}
        \toprule
       \textbf{Model}  &  \textbf{Het.} & \textbf{Scale} &\textbf{LRI}& \textbf{Dec.} & \textbf{Multi-Target}\\
       \midrule
       Dec. MPC\cite{tallamraju2018decentralized} &  $\times$ & -  & $\times$ & \checkmark & $\times$ \\
       PSO \cite{oh2016pso}&  $\times$ & -  && $\times$ & \checkmark \\
       GNN~\cite{li2020graph} &  $\times$ & \checkmark && \checkmark & $\times$ \\
       GNN~\cite{zhou2022graph} &  $\times$ & \checkmark& $\times$ &\checkmark & $\times$ \\
       HetNET \cite{seraj2022learning}& \checkmark & \checkmark  & \checkmark & $\times$ & - \\
       HetGPPO \cite{hetGPPO}&\checkmark&-&-&\checkmark&-\\
       MAGAT\cite{li2021message} &  $\times$ & \checkmark  &$\times$& \checkmark & $\times$ \\
       \textbf{GATAR} &  \checkmark & \checkmark &\checkmark& \checkmark & \checkmark \\
        \bottomrule
    \end{tabular}
    }
    \caption{Analyzing various models for their ability to incorporate heterogeneity (Het.), scalability (Scale), long-range interaction (LRI),  decentralization (Dec.) and whether they account for multiple target tracking}
    \vspace{-12pt}
    \label{tab:model-eval}
\end{table}

\textbf{Generalizability to Robot Configurations}: To examine the generalizability of our methods on team compositions, we tested our model on unseen team composition and team size, up to 6 air vehicles and 6 ground vehicles. We test on a team size ranging from two to twelve robots, with a total of six kinds of team composition. 

\textbf{Baseline Comparisons}: Our method is compared with the greedy algorithm without communication, in which robots will simply choose the nearest targets as their tasks, Graph Attention Networks (GAT) and Graph Neural Network (GNN) based models, and a random selection, in which a random target in the field of view is selected. To the best of our knowledge, we are the first to solve the decentralized multi-robot multi-targets target localization problem under a heterogeneous setting, as shown in Table \ref{tab:model-eval}. As shown in Fig., \ref{fig:teamsize}, \method ~ achieve at most a 43.2\% decrease in distance error compared to greedy without communication. At the same time, \method ~consistently outperforms baseline methods, showing that it uses information from neighbors and has a more accurate prediction on a single preference. Table \ref{tab:baselines} compares the performance of GATAR against other baseline neural network models. In this part, all the learnable methods are trained for 20 epochs. It can be seen that GATAR outperforms all the other models.  
\begin{table}[!th]
\centering
\begin{tabular}{llll}
\toprule
\textbf{Method} & \textbf{90-percentile dist.} & \textbf{median dist.} & \textbf{mean dist.} \\ \midrule
GATAR & \textbf{4.3928} & \textbf{0.8156} & \textbf{1.6308} \\
GNN \cite{zhou2022graph} & 4.5618 & 0.8755 & 1.7180 \\
GNN+V-Cycle & 4.6152 & 0.8452 & 1.6996 \\
MAGAT \cite{li2021message} & 5.4553 & 0.8588 & 1.8746 \\
\bottomrule
\end{tabular}
\caption{Comparison of GATAR performance against baseline neural network-based architectures, on the 6A6G setting}
\vspace{-10pt}
\label{tab:baselines}
\end{table}

\begin{table}[h]
\centering
\resizebox{0.45\textwidth}{!}{
\begin{tabular}{llllll}
\toprule
\multicolumn{3}{c}{\textbf{Het-Aware Preprocessing}} &
  \multicolumn{1}{c}{\multirow{2}{*}{\textbf{V-Cycle}}} &
  \multirow{2}{*}{\textbf{median dist.}} &
  \multirow{2}{*}{\textbf{mean dist.}} \\ 
\multicolumn{1}{c}{\makebox[0.02\textwidth]GT\makebox[0.02\textwidth]} &
  \multicolumn{1}{c}{\makebox[0.02\textwidth]PM\makebox[0.02\textwidth]} &
  \multicolumn{1}{c}{\makebox[0.02\textwidth]RM\makebox[0.02\textwidth]} &
  \multicolumn{1}{c}{} &
  \multicolumn{1}{c}{} &
   \\ \midrule
\multicolumn{1}{l}{-} & \multicolumn{1}{l}{-} & - & \checkmark & 1.47 & 2.02 \\ 
\multicolumn{1}{l}{\checkmark} & \multicolumn{1}{l}{\checkmark} & - & \checkmark & 1.41  & 2.00  \\ 
\multicolumn{1}{l}{\checkmark} & \multicolumn{1}{l}{-} & \checkmark & \checkmark & 1.19   &  1.86 \\ 
\multicolumn{1}{l}{\checkmark} & \multicolumn{1}{l}{\checkmark} & \checkmark & - & 1.59   & 2.13 \\ 
\multicolumn{1}{l}{\checkmark} & \multicolumn{1}{l}{\checkmark} & \checkmark & \checkmark & \textbf{0.81}   & \textbf{1.63} \\ \bottomrule
\end{tabular}}
\caption{Ablation studies; \textbf{GT} means \textit{Gaussian targets map}; \textbf{PM} refers to the product between targets map and reward map; \textbf{RM} means the \textit{Reward map}. \textbf{50th}  and \textbf{avg} means the 50th percentile distance and average distance.}
\vspace{-18pt}
\label{tb:ablation}
\end{table}
~

\vspace{-10pt}
\textbf{Effectiveness of Heterogeneity-Aware Preprocessing}: Shown in Table. \ref{tb:ablation}, we tested the effectiveness of different components in Heterogeneity-Aware Preprocessing.  Rows 1,3, and 5 show that the Gaussian targets map and Reward Map improve the performance mainly in general cases, as the 50th percentile distance sees a more significant decrease when these two components are included. Rows 2 and 5 show that the reward map is beneficial both for corner cases and general cases, as its removal will lead to a significant performance drop on both \texttt{dist-50} and \texttt{dist-avg}. Pixel Wise product helps with the overall performance when the other two metrics are included but reduces the accuracy in the absence of either of the other two metrics.

\textbf{Effectiveness of V-Cycle Structure}: We also tested the effectiveness of the V-Cycle structure. By comparing rows 4 and 5, in its absence, both median distance and mean distance will have a significant increase, indicating that this structure helps the model aggregate information from long-range neighbors efficiently.

\vspace{-11pt}
\subsection{Results on Environments defined on Google Maps data}\label{subsec:qual}

\begin{figure}[!h]
\vspace{-5pt}
    \centering
    \begin{subfigure}[b]{0.2\textwidth}
        \centering
        \includegraphics[width=\textwidth]{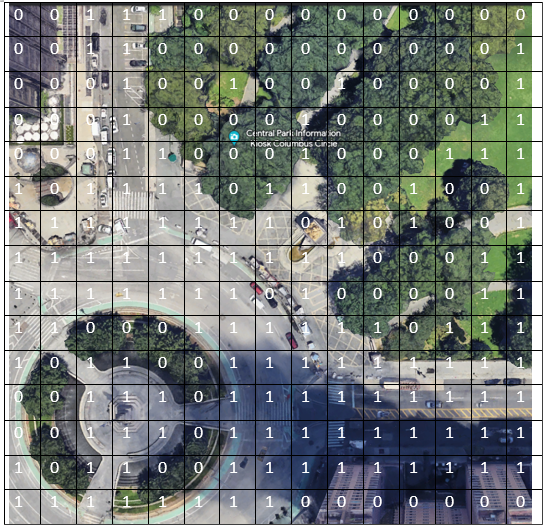}
        \caption{Original Map}
        \label{fig:realworld}
    \end{subfigure}
    \begin{subfigure}[b]{0.2\textwidth}
        \centering
        \includegraphics[width=\textwidth]{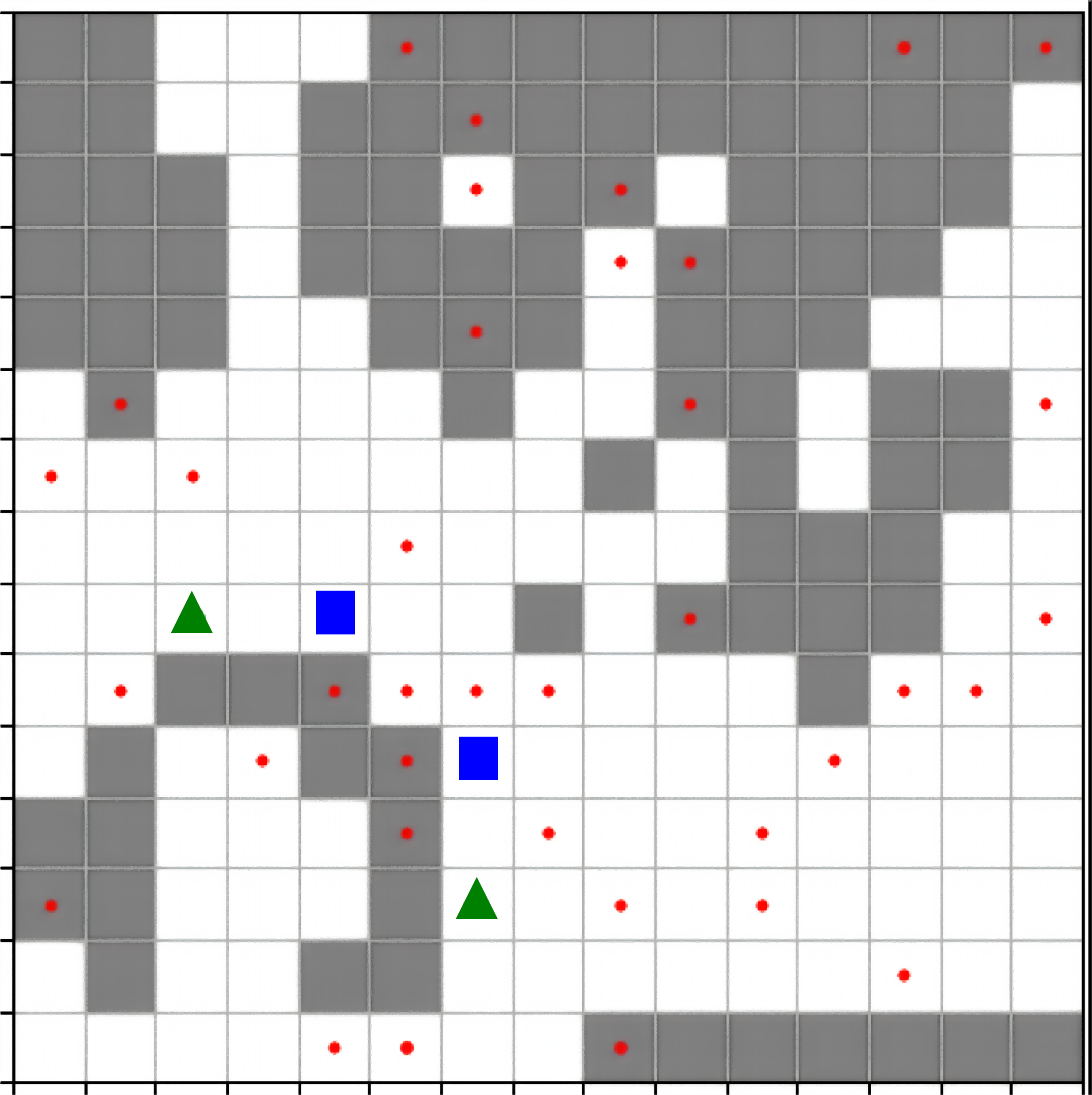}
        \caption{Environment}
        \label{fig:gridw}
    \end{subfigure}
    \begin{subfigure}[b]{0.2\textwidth}
        \centering
        \includegraphics[width=\textwidth]{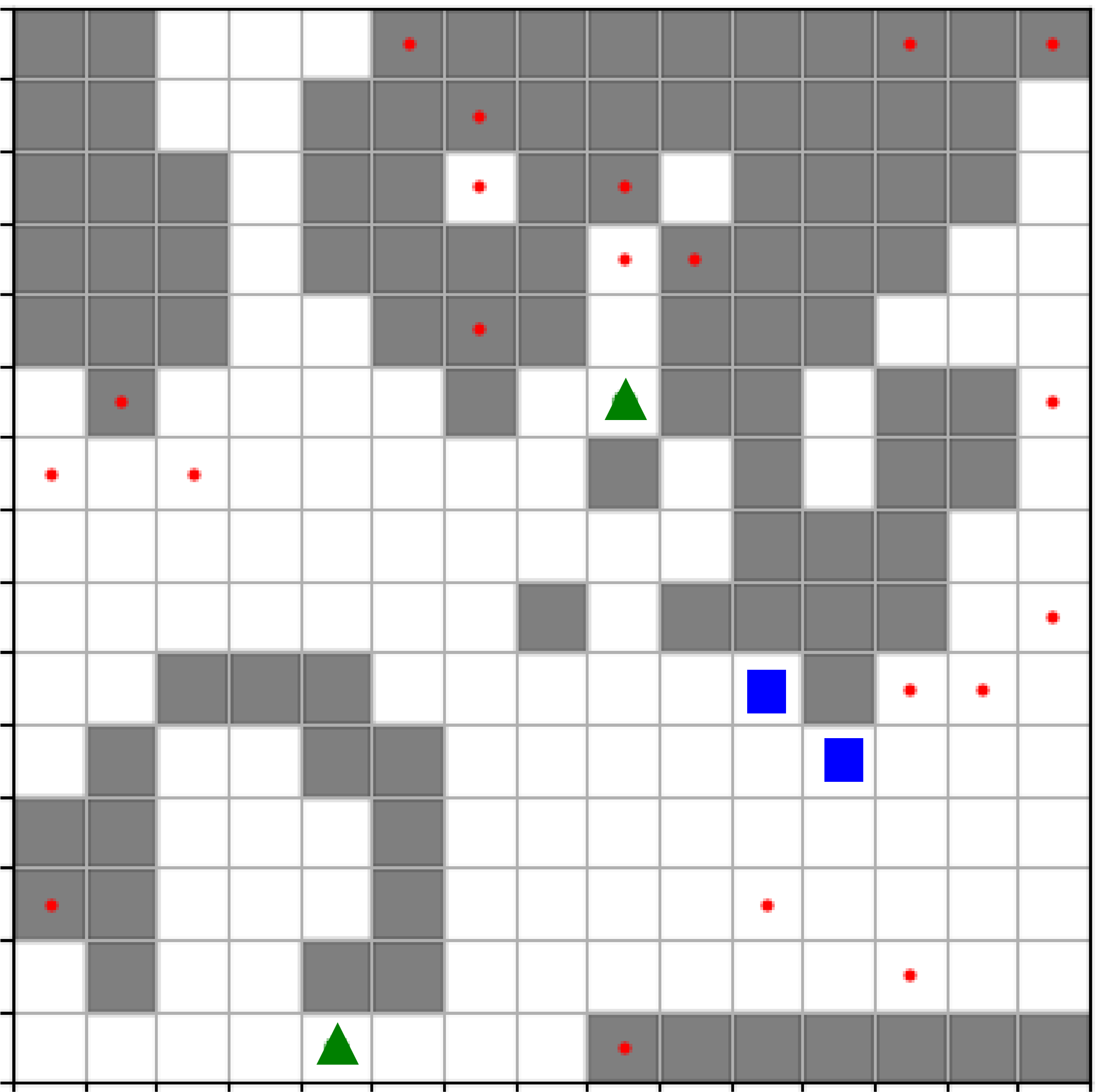}
        \caption{Running State}
        \label{fig:ongoing}
    \end{subfigure}
    \begin{subfigure}[b]{0.2\textwidth}
        \centering
        \includegraphics[width=\textwidth]{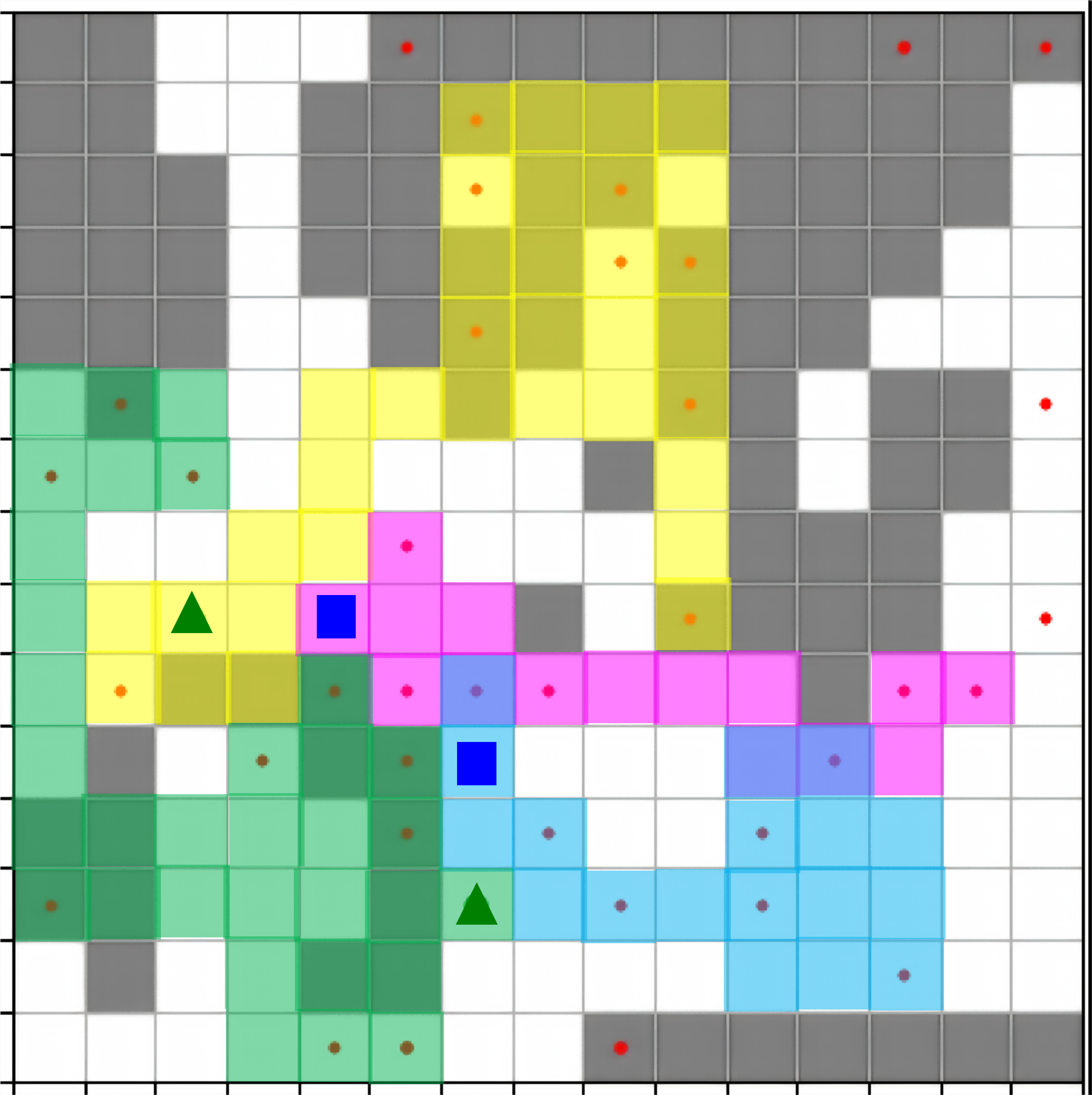}
        \caption{Targets Coverage}
        \label{fig:coverage}
    \end{subfigure}
    \caption{\textbf{Demostration of task allocation in a real-world scenario.} The map is a birds-eye-view image of the Columbus Circle by Central Park in Mid-town Manhattan, New York City. \textcolor{red}{Red} points indicate the randomly distributed targets; \textcolor{blue}{Blue}  squares and  \textcolor{darkgreen}{Green}  triangles indicate UGVs and UAVs initially located around the roundabout; Grey squares indicate the obstacles such as buildings and shrubs.}
    \vspace{-20pt}
    \label{fig:quali}
\end{figure}
To evaluate the performance on real-world-like scenarios, we generated an experimental environment (Fig.\ref{fig:gridw}) from Google Maps snapshots, as shown in (Fig.\ref{fig:realworld}) and performed the sequential task allocation on it. Fig. \ref{fig:ongoing} is a chosen frame from the whole process, showing that the UAVs are exploring obstacle-rich areas while the UGVs are exploring obstacle-free space. Fig. \ref{fig:coverage} shows the targets covered by different robots within 40 steps. When compared with the actual map, the UAVs cover areas around Central Park and roundabouts with dense vegetation, while the UGVs operate on broad streets and sidewalks.
\vspace{-10pt}

\section{Conclusion}
\vspace{-3.5pt}
In this paper, we present \method, a decentralized method for task allocation in a target localization scenario.
We show that \method ~ can operate on different types of robots using a single scalable model by utilizing heterogeneity-aware feature preprocessing. 
Comprehensive experiments show that with the V-Cycle structure and heterogeneity-aware preprocessing, our method has near-optimal performance and good scalability on team size and composition.Limitations still occur in the connection to upstream perception and downstream planning. In future work, we aim to extend our method to an end-to-end system and deploy it on real-world multi-robot systems. We will also explore the model's robustness against latency in multi-round aggregation and the efficient use of historical collaborative data. 
\vspace{-10pt}

\section*{Acknowledgment}
We would like to thank Dr. Kshitij Tiwari for his valuable input regarding heterogeneous multi-agent systems and their applications.

\vspace{-10pt}

\bibliographystyle{IEEEtran}
\bibliography{root}

\end{document}